\documentclass[10pt,twocolumn,letterpaper]{article}

\usepackage{iccv}
\usepackage{times}
\usepackage{epsfig}
\usepackage{graphicx}
\usepackage{amsmath}
\usepackage{amssymb}
\usepackage{color}
\usepackage{ulem}
\usepackage{ifthen}
\usepackage{algorithm}
\usepackage[noend]{algpseudocode}
\usepackage{array,multirow}
\usepackage{adjustbox}

\definecolor{gray}{rgb}{0.5,0.5,0.5} 
\definecolor{green}{rgb}{0, 0.4, 0} 
\definecolor{orange}{rgb}{1, 0.5, 0} 	
\definecolor{mahogany}{rgb}{0.75, 0.25, 0.0}
\definecolor{purple}{rgb}{0.6, 0, 0.6}
\definecolor{purple}{rgb}{0.6, 0, 0.6}
\definecolor{darkgreen}{rgb}{0, 0.4, 0} 
\definecolor{frenchblue}{rgb}{0.0, 0.45, 0.73}

\newboolean{revising}
\setboolean{revising}{False}
\ifthenelse{\boolean{revising}}
{
	\newcommand{\ignore}[1]{}

	\newcommand{\gina}[1]{\textcolor{darkgreen}{#1}}

	\newcommand{\chienreplace}[2]{\textcolor{orange}{#2}}
	\newcommand{\sunmin}[1]{\textcolor{purple}{#1}}
	
    \newcommand{\highlight}[1]{\textcolor{red}{#1}}
} {
	\newcommand{\ignore}[1]{}

	\newcommand{\gina}[1]{#1}

	\newcommand{\chienreplace}[2]{#2}
	\newcommand{\sunmin}[1]{#1}
	
	\newcommand{\highlight}[1]{#1}
}

\newcommand{\cutsectionup}{\vspace*{-0.08in}}%{{\vspace*{-0.03in}}}
\newcommand{\cutsectiondown}{\vspace*{-0.08in}}%{{\vspace*{-0.01in}}}

\newcommand{\cutsubsectionup}{\vspace*{-0.1in}} %{\vspace*{-0.1in}}}
\newcommand{\cutsubsectiondown}{\vspace*{-0.1in}} %{\vspace*{-0.04in}}}

\newcommand{\cutfigureup}{\vspace*{-0.15in}}%{{\vspace*{-0.1in}}}
\newcommand{\cutfiguredown}{\vspace*{-0.15in}}%{{\vspace*{-0.1in}}}

\usepackage[breaklinks=true,bookmarks=false]{hyperref}

\iccvfinalcopy % *** Uncomment this line for the final submission

\def\iccvPaperID{316} % *** Enter the ICCV Paper ID here

\ificcvfinal\fi
\begin{document}

%%%%%%%%% TITLE
\title{Anticipating Daily Intention using On-Wrist Motion Triggered Sensing}

\author{Tz-Ying Wu\footnotemark[1], Ting-An Chien\thanks{indicates equal contribution}, Cheng-Sheng Chan, Chan-Wei Hu, Min Sun \\
Dept. of Electrical Engineering, \\
National Tsing Hua University, Taiwan.\\
{\tt\small \{gina9726, tingan0206, james121506, huchanwei1204, aliensunmin\}@gmail.com}
%\and
}

\maketitle
\thispagestyle{empty}

%%%%%%%%% ABSTRACT
\begin{abstract}
   Anticipating human intention by observing one's actions has many applications. For instance, picking up a cellphone, then a charger (actions) implies that one wants to charge the cellphone (intention) (Fig.~\ref{fig.teaser}). By anticipating the intention, an intelligent system can guide the user to the closest power outlet.
We propose an on-wrist motion triggered sensing system for anticipating daily intentions, where the on-wrist sensors help us to persistently observe one's actions.
The core of the system is a novel Recurrent Neural Network (RNN) and Policy Network (PN), where the RNN encodes visual and motion observation to anticipate intention, and the PN parsimoniously triggers the process of visual observation to reduce computation requirement. We jointly trained the whole network using policy gradient and cross-entropy loss.
To evaluate, we collect the first daily ``intention" dataset consisting of $2379$ videos with $34$ intentions and $164$ unique action sequences (paths in Fig.~\ref{fig.teaser}). Our method achieves $92.68\%,90.85\%,97.56\%$ accuracy on three users while processing only $29\%$ of the visual observation on average.
\end{abstract}

%%%%%%%%% BODY TEXT
\section{Introduction}\label{sec.Intro}

\begin{figure}[t!]
\cutfigureup
\begin{center}
\includegraphics[width=0.975\linewidth]{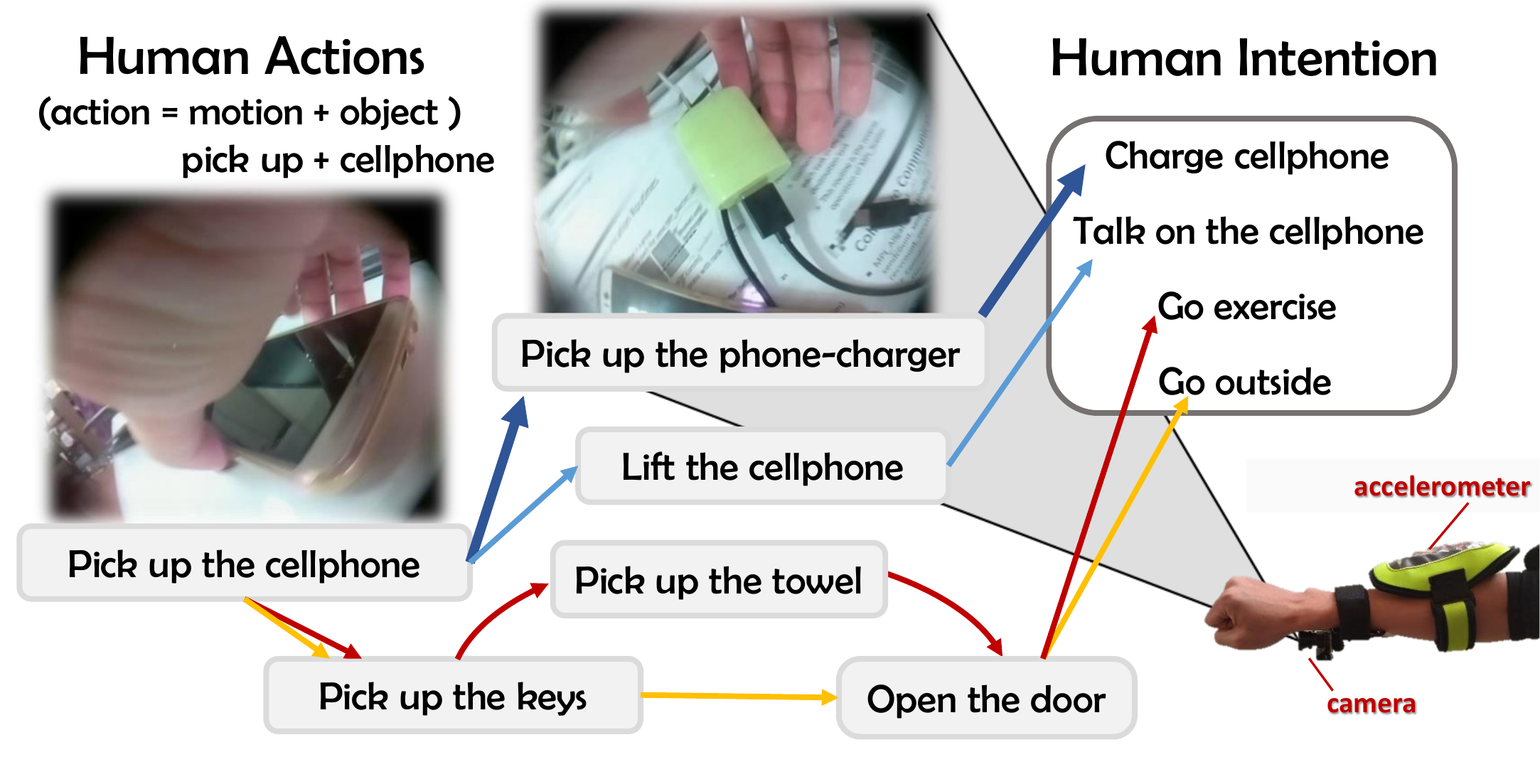}
\end{center}
\vspace{-6mm}
\caption{\small Illustration of intention anticipation. An action sequence (i.e., an ordered list of actions) is a strong cue to anticipate intention -- predicting an intention before it occurs. For instance, the actions on the dark blue path (i.e., pick-up the cellphone; then, pick-up cellphone charger) strongly imply ``charge cellphone". The task is challenging since (1) the same action (pick-up the cellphone) can lead to different intentions (talk on the cellphone vs. charge cellphone), and (2) multiple paths can lead to the same intention (see Fig.~\ref{fig.ID}). Bottom-right panel: actions are recorded by our on-wrist sensors including a camera and an accelerometer.}\label{fig.teaser}
\cutfiguredown
\end{figure}

% intelligent system
Thanks to the advance in Artificial Intelligence, many intelligent systems (e.g., Amazon Echo, Google Home.) have become available on the markets. Despite their great ability to interact with humans through a speech interface, they are currently not good at proactively interacting with humans.
Thus, we argue that the key for proactive interaction is to anticipate user's intention by observing their actions. Given the anticipated intention, the intelligent system may provide service to facilitate the intention. More specifically, the ability to anticipate a large number of daily intentions will be the key to enable a proactive intelligent system.

% anticipate intention
Many researchers have tackled tasks related to intention anticipation. \cite{Hoai-DelaTorre-CVPR12,Ryoo11,Lan2014} focus on early activity prediction -- predicting actions before they have completed. However, the time-to-action-completion of this task is typically very short. Hence, there are only a few scenarios that intelligent systems may take advantage of the predicted activity.
Kitani et al.~\cite{Kitani_2012_7250} propose to forecast human's trajectory. Forecasting trajectory is very useful, but it does not directly tell you the ``intention" behind a trajectory. \cite{ACCV_acident,JainICCV15,JainICRA16} anticipate the future events on the road such as making a left turn or involving in an accident. Although these events can be considered as intentions, only few intentions (at most five) are studied. Moreover, none of the work above leverages heterogeneous sensing modalities to reduce computing requirement.
%We argue that the ability to anticipate a much larger number of daily intention will be required to enable a proactive intelligent system.

% out task in smart home
In this work, we anticipate a variety of daily intentions (e.g., ``go outside", ``charge cellphone", in Fig.~\ref{fig.teaser}) by sensing motion and visual observation of actions.
Our method is unique in several ways. Firstly, we focus on \textit{On-Wrist} sensing: (1) an on-wrist camera (inspired by \cite{Ohnishi_2016_CVPR,ChanECCV16}) is used to observe object interactions reliably, and (2) an on-wrist accelerometer is used to sense 3D hand motion efficiently.
Since both on-wrist sensors are unconventional, we collect auxiliary object appearance and motion data to pre-train two encoders: (1) a Convolutional Neural Network (CNN) to classify daily objects, and (2) a 1D-CNN to classify common motions.
%We show that pretraining consistently improve the accuracy of our system.
%Intuitively, a unique pair of motion (e.g., pick up) and object (e.g., cellphone) corresponds to an action (e.g., pick up cellphone). We prove that object and motion are complementary since our model achieves the best performance by sensor fusion.
Secondly, we leverage heterogeneous sensing modalities to reduce computing requirement. Note that visual data is very informative but costly to compute. In contrast, motion data is less informative but cheap to compute. 
We propose a Policy Network to determine when to peek at some images. The network will trigger the camera only at some important moments while continuously analyzing the motions. We call this as \textit{Motion Triggered} sensing.
Finally, we propose to use a Recurrent Neural Network (RNN) to model important long- and short-term dependency of actions. Modeling this dependency properly is the key of accurate anticipation, since daily action sequences are subtle and diverse. For instance, while multiple action sequences leading to the same intention, the same subset of actions can lead to different intention as well (see ``go exercise" and ``go outside" in  Fig.~\ref{fig.teaser}).
% Our RNN model also robustly fuses object and motion observations through time.}

%\chan{Besides,} it is hard to collect daily intention datasets due to privacy concern. In contrast, separately recorded objects and motions datasets, which do not involve human intention, have less privacy concern.
%Hence, we pre-train two separate encoders: (1) a Convolutional Neural Network (CNN) to classify common daily objects, and (2) a 1D-CNN to classify common motions. %We show that object and motion pre-training are very effective on improving intention anticipation performance.
%Daily intention anticipation is challenging, since there are multiple sequences of actions leading to the same intention.
%For instance, \hl{before going outside, a user may either picks up a car key before picking up backpack, or the other way around.} However, in some cases, the order of action is important.
%When a large number of actions is required to fulfill an intention, there exist a large number of valid sequence of actions to fulfill the intention.
%To the best of our knowledge, none of the previous work addresses all these challenges in daily intention anticipation.

In order to evaluate our method, we collect the first daily intention dataset from on-wrist sensors. It consists of 2379 videos with 34 intentions and 164 unique action sequences. For pre-training encoders, we collect an object dataset by manipulating 50\footnote{including a hand free class, which means that hand is not interacting with any objects.} daily objects without any specific intention, and a 3D hand motion dataset with six motions performed by eight users. 
On the intention dataset, our method achieves $92.68\%,90.85\%,97.56\%$ accuracy while processing only $29\%$ of the visual observation on average.

Our main contributions can be summarized as follows.
%In our experiment, we show that both pre-trained CNN-encoders and sensor fusion improve intention anticipation performance. 
%\begin{itemize}
(1) We adapt on-wrist sensors to reliably capture daily human actions.
%\sout{(2) We show that fusing object and motion sensors effectively improve intention anticipation performance.}
%\item We jointly anticipate intention and explanatory sequence of actions such that users can interactively improve performance.
%(2) We show that pre-training object and motion encoders using auxiliary data with less privacy concern is effective.
(2) We show that our policy network can effectively select the important images while only slightly sacrificing the anticipation accuracy.
(3) We collected and will release one of the first daily intention dataset with a diverse set of action sequence and heterogeneous on-wrist sensory observations.
%\end{itemize}

% behavior from object interaction and motion

% applications
\section{Related Work}\label{sec.RW}

%\highlight{RW is copied from CVPR paper, unchanged!}

We first describe works related to anticipation. Then, we mention other behavior analysis tasks. Finally, we describe a few works using wearable sensors for recognition.

\subsection{Anticipation}

The gist of anticipation is to predict in the future. We describe related works into groups as follows.

\noindent\textbf{Early activity recognition.}
\cite{Hoai-DelaTorre-CVPR12,Ryoo11,Lan2014} focus on predicting activities before they are completed. For instance, recognizing a smile as early as the corners of the mouth curve up.
Ryoo~\cite{Ryoo11} introduces a probability model for early activity prediction.
Hoai et al.~\cite{Hoai-DelaTorre-CVPR12} proposed a max-margin model to handle partial observation. Lan et al.~\cite{Lan2014} propose the hierarchical movemes representation for predicting future activities.

\noindent\textbf{Event anticipation.}
 \cite{koppula2016anticipating,JainICRA16,JainICCV15,ACCV_acident,Vondrick_2016_CVPR} anticipate events even before they appear.
Jain et al.~\cite{JainICRA16,JainICCV15} propose to fuse multiple visual sensors to anticipate the actions of a driver such as turning left or right. Fu et al.~\cite{ACCV_acident} further propose a dynamic soft-attention-based RNN model to anticipate accidents on the road captured in dashcam videos.
Recently, Vondrick et al.~\cite{Vondrick_2016_CVPR} propose to learn temporal knowledge from unlabeled videos for anticipating actions and objects.
However, the early action recognition and anticipation approaches focus on activity categories and do not study risk assessment of objects and regions in videos.
Bokhari and Kitani~\cite{KACCV16} propose to forecast long-term activities from a first-person perspective.

\noindent\textbf{Intention anticipation.}
Intention has been explored more in the robotic community~\cite{WangDBVSP2012,koppula2016anticipating,Koppula2016,BerensonIROS13}.
Wang et al.~\cite{WangDBVSP2012} propose a latent variable model for inferring human intentions.
Koppula and Saxena~\cite{koppula2016anticipating} address
the problem by observing RGB-D data. A real robotic system has executed the proposed method to assist humans in daily tasks.
\cite{Koppula2016,BerensonIROS13} also propose to anticipate human activities for improving human-robot collaboration.
Hashimoto et al.~\cite{Intention2016} recently propose to sense intention in cooking tasks via the knowledge of access to objects. \sunmin{Recently, Rhinehart and Kitani~\cite{DARKO} propose an on-line approach for first-person videos to anticipate intentions including where to go and what to acquire.}

\noindent\textbf{Others.}
Kitani et al. \cite{Kitani_2012_7250} propose to forecast human trajectory by surrounding physical environment (e.g., road, pavement).
The paper shows that the forecasted trajectory can be used to improve object tracking accuracy.
%\sunmin{Bokhari and Kitani~\cite{KACCV16} propose to forecast both human trajectory and a discrete sequence of sub-actions for long-term activity forecasting.}
Yuen and Torralba~\cite{Yuen:2010} propose to predict motion from still images.
Julian et al. \cite{walker2014patch} propose a novel visual appearance prediction method based on mid-level visual elements with temporal modeling methods.

Despite many related works, to the best of our knowledge, this is the first work in computer vision focusing on leveraging a heterogeneous sensing system to anticipate daily intentions with low computation requirement.

\subsection{High-level Behavior Analysis}

Other than activity recognition, there are a few high-level behavior analysis tasks. Joo et al.~\cite{Joo_2014_CVPR} propose to predict the persuasive motivation of the
photographer who captured an image. Vondrick et al.~ \cite{Vondrick_2016_CVPR} propose to infer the motivation of actions in an image by leveraging text. Recently, many methods (e.g., \cite{Paragraph,PanXYWZ15,PanMYLR15,HaoTitleECCV16,S2SICCV15,SAICCV15}) have been proposed to generate sentence or paragraph to describe the behavior of humans in a video. %Despite these tasks are novel, it is less clear how to practically 

\begin{figure*}[t!]
\cutfigureup
\begin{center}
\includegraphics[width=0.975\linewidth]{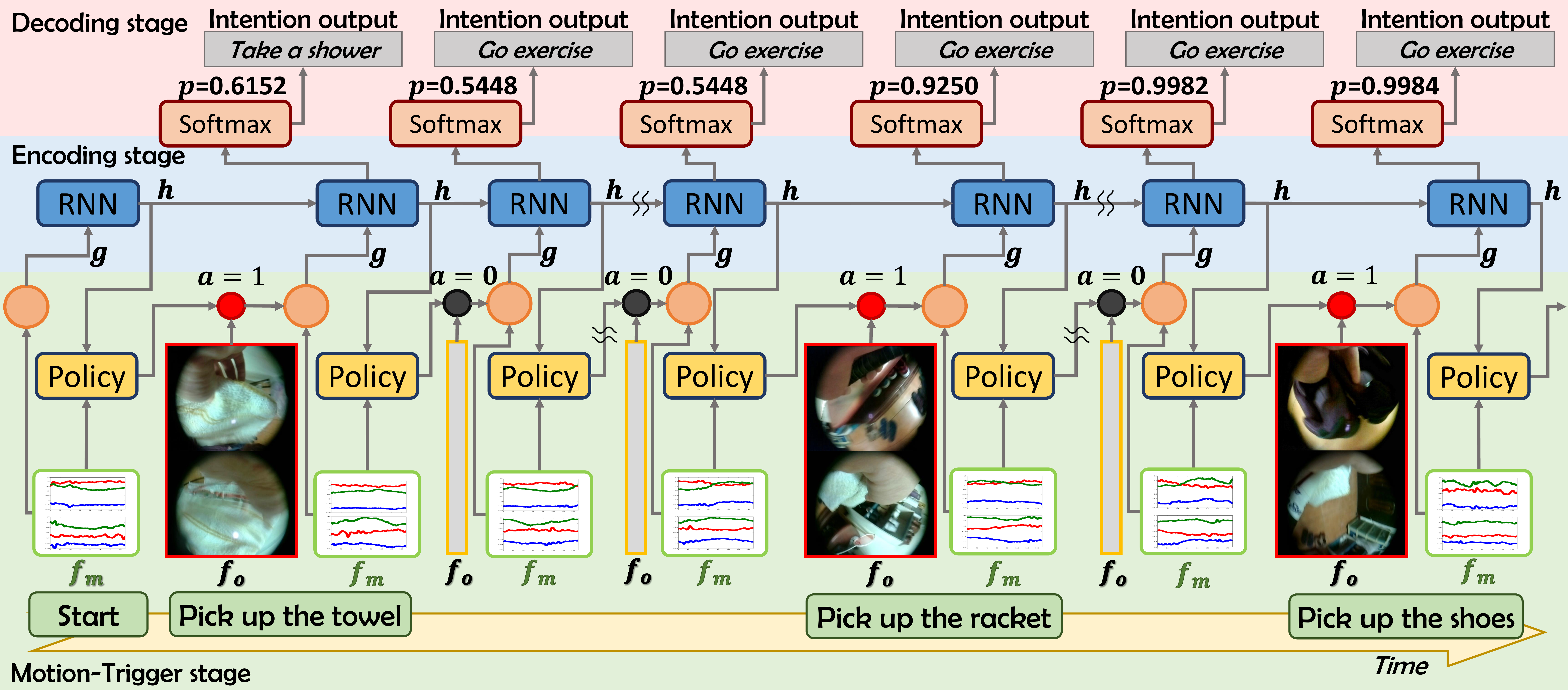}
\end{center}
\vspace{-4mm}
\caption{\small Visualization of our motion-triggered model. Our model consists of an RNN with LSTM cell encoder (blue block) and a Policy Network (yellow block). At each frame, RNN will generate an anticipated intention according to a new embedded representation $g$ and the previous hidden state $h$ of the RNN. The policy will generate the motion-trigger decision $a$ for next frame, based on motion representation $f_m$ and the hidden state $h$ of the RNN.
The orange circle represents the fusion operation (details in Sec.~\ref{sec.model}). The red and black circles represent a trigger and non-trigger decision of policy network, respectively (details in Sec.~\ref{sec.Policy}). When $a=0$, $f_o$ is empty since it is not processed.}
\label{fig.model}
\cutfiguredown
\end{figure*}

%to observe a sequence of objects and motions. At each frame, our model has two fusion layers: (1) object and motion fusion at each hand (fusing $f$ from object $o$ and motion $m$), and (2) left (green panel) and right (yellow panel) hands fusion (orange circle). See more details in Sec.~\ref{sec.model}.
%This example shows anticipation given $100\%$ observation. 
%Our model can anticipate intention $p$ from the RNN at an arbitrary time. We show the maximum anticipation probability becomes higher when observing more video frames ($0.4257$ to $0.9999$).

\subsection{Recognition from Wearable Sensors}

Most wearable sensors used in computer vision are first-person (i.e., ego-centric) cameras. \cite{ma2016going,Singh_2016_CVPR,fathi2011understanding,li2015delving} propose to recognize activities. \cite{lu2013story,ghosh2012discovering} propose to summarize daily activities.
Recently, two works \cite{Ohnishi_2016_CVPR,ChanECCV16} focus on recognition using on-wrist camera and show that it outperforms ego-centric cameras. Inspired by them, we adapt a similar on-wrist sensor approach.

%\subsection{Daily Activity Datasets}
%Daily activity data is hard to collected at large scale, since there are privacy concerns. 

\section{Our Approach}\label{sec.Tech}

We first define the problem of intention anticipation.
Next, we introduce our RNN model encoding sequential observations and fusing multiple sensors' information from both hands. Then, we talk about our novel motion-triggered process based on a policy network. Finally, we describe how we pre-train the representation from auxiliary data.

\subsection{Problem Formulation}\label{sec.form}

%We define the intention anticipation task below.

\noindent\textbf{Observations.}
At frame $t$, the camera observes an image $I_t$, and the motion sensor observes the 3D acceleration of hands $A_t\in R^3$.

\noindent\textbf{Representations.}
The image $I$ and 3D acceleration $A$ are raw sensory values which are challenging to be used directly for intention anticipation, especially when lacking training data.
Hence, we propose to learn visual object (referred to as object) $f_{o,t}$ and hand motion (referred to as motion) $f_{m,t}$ representations from other tasks with a larger number of training data.
%interaction (referred to as object for simplicity)
Note that, for all the variables, we use superscript to specify left or right hand (when needed). For instance, $f_{o,t}^L$ indicates left-hand object representation.
%at frame $t$.
%since our approach has observations from both left and right hands

\noindent\textbf{Goal.}
At frame $t$, our model predicts the future intention $y_t\in\mathcal{Y}$ based on the observations, where $\mathcal{Y}$ is the set of intention indices. Assuming the intention occurs at frame $T$,
we not only want the prediction to be correct but also to predict as early as possible (i.e., $T-t$ to be large).

% N
% In order to train an anticipation model, we follow \cite{JainICCV15} to use the following exponential Loss,
% \begin{eqnarray}
% L_t(p_t,y_t^*) = -e^{-(T-t)}\log(p_t(y_t^*)),\label{eq.exp}
% \end{eqnarray}
% where $T$ is the frame when intention occurs and $y_t^*$ is the ground truth intention.
% The exponential loss panelizes more if intention is not predicted correctly at time closer to $T$.
% Since the whole system differentiable,
% we train our model supervisedly using backpropagation.

\subsection{Our Recurrent and Fusion Model}\label{sec.model}
Intention anticipation is a very challenging task. Intuitively, the order of observed objects and hand motions should be a very strong cue. However, most orders are not strict. Hence, learning composite orders from limited training data is critical. 

\noindent\textbf{Recurrent Neural Network (RNN) for encoding.}
We propose to use an RNN with two-layers of Long-Short-Term-Memory (LSTM) cell to handle the variation (Fig.~\ref{fig.model}-Top) as follows,
%\begin{eqnarray}
\begin{align}
%p_t=\textrm{softmax}(W_y\textrm{ } h_t),\\
&g_t=\textrm{Emb}(W_{emb}, \textrm{con}(f_{m,t}, f_{o,t}))~,&\label{eq.g}\\
&h_t=\textrm{RNN}(g_t, h_{t-1})~,&\\
&p_t=\textrm{Softmax}(W_y, h_t)~,&\label{eq1}\\
&y_t=\arg\max_{y\in\mathcal{Y}} p_t(y)~,&
\end{align}
%\end{eqnarray}
where $p_t\in R^{|\mathcal{Y}|}$ is the softmax probability of every intention in $\mathcal{Y}$, $W_y$ is the model parameter to be learned,
%\chanreplace{$\textrm{softmax}$ is the softmax function ensuring $p_t$ is a valid probability,}{}
$h_t$ is the learned hidden representation, and $g_t$ is a fixed dimension output of $\textrm{Emb}(\cdot)$. $W_{emb}$ is the parameter of embedding function $\textrm{Emb}(\cdot)$, $\textrm{con}(\cdot)$ is the concatenation operation, and $\textrm{Emb}(\cdot)$ is a linear mapping function (i.e., $g=W_{emb}\cdot \textrm{con}(f_m,f_o,1)$.
%Note that $p_t(y_t)$ is the probability that intention $y_t$ is predicted.
RNN has the advantage of learning both long- and short-term dependency of observation which is ideal for anticipating intentions.

%\chanreplace{Recall that our system has sensors on both left and right hands. We first describe how to fuse camera and motion sensors; then, we describe how to fuse left and right hand observations.} {Recall that our system has two types of sensors, camera and accelerometer, on both left and right hands. We describe how to fuse all of these features as follows.}

%\noindent\textbf{Fusing object and motion.}
%We first separately normalize the feature representations of both object $f_o$ and motion $f_m$ to range from $-1$ to $1$. \chan{Then, we simply concatenate both representations as the fused representation. $f_{o+m}=\textrm{con}(f_o,f_m)$, where $\textrm{con}$ denotes concatenation.}

\noindent\textbf{Fusing left and right hands.}
Since tasks in real life typically are not completed by only one hand, we allow our system to observe actions on both hands simultaneously.
We concatenate the right (i.e., the dominant hand) and left-hand observations in a fixed order to preserve the information of which hand is used for certain actions more frequently. The fused observation is $f_{i} = \textrm{con}(f_{i}^R,f_{i}^L)$, where $i \in\{o, m\}$.
%In addition, the dimension of feature $f$ \ginareplace{will change from motion to object.}{is varying since $f$ can be $f_m$ or $f_{o+m}$.} Hence, there is a embedding function defined as $Emb$: $f \rightarrow g$. 
%The dimension of $g$ is fixed, but the dimension of $f$ can be changed, where $f \in \{f_o, f_m, f_{o+m}\}$.
%Note that we concatenate the right hand with left hand by assuming right hand as the dominant hand. \\
%\gina{(Q: Does this imply the dominant hand should be placed first? may be asked by RW)} 

%\noindent\textbf{Sequence encoding using LSTM}
%\subsection{Training}
%\noindent\textbf{Training with exponential loss.}
\noindent\textbf{Training for anticipation.}
Since our goal is to predict at any time before the intention happened, anticipation error at different time should be panelized differently.
We use exponential loss to train our RNN-based model similar to \cite{JainICCV15}. The anticipation loss $L^A$ is defined as,
%\begin{eqnarray}
\begin{align}
\sum_{t=1}^{T} L^A_t=\sum_{t=1}^{T}-\log p_t(y^{\textrm{gt}})\cdot e^{\log(0.1) \frac{T-t}{T}}~,\label{eq.exploss}
\end{align}
%\end{eqnarray}
where $y^{\textrm{gt}}$ is the ground truth intention and $T$ is the time when intention reached. Based on this definition, the loss at the first frame (t=0) is only $10\%$ of last frame (t=T). This 
implies that anticipation error is panelized less when it is early, but more when it is late.
This encourages our model to anticipate the correct intention as early as possible.

The current RNN considers both motion $f_m$ and object $f_o$ representations as shown in Eq.~\ref{eq.g}. It is also straightforward to modify Eq.~\ref{eq.g} such that RNN considers only motion or only object representation. However, the RNN needs to consider the same type of representation at all times. In the following section, we introduce the Motion-Triggered sensing process, where the RNN considers different representations at different frames depending on a learned policy.

%\chan{Based on this architecture, the RNN encoder can be trained for encoding $f_o, f_m,$ and $f_{o+m}$. It didn't involve the Motion-Triggered process at this stage, the RNN model will directly anticipate the intentions by sensing complete observations. We'll evaluate and report the results in Sec.xx.}
%we simply train our model to make correct intention prediction given $100\%$ observations. This is because we found that training with exponential loss in our case gives rise to very unstable training loss behaviors. One major difference between our anticipation task and the tasks in \cite{JainICCV15,JainICRA16,ACCV_acident} is that the duration (in seconds) of action sequences \chanreplace{for different daily intentions are significantly different.}{for different daily intentions.} Both our empirical results and observation suggest that the original exponential loss cannot be directly applied.

\subsection{RL-based Policy Network}\label{sec.Policy}
We propose a policy network $\pi$ to determine when to process a raw image observation $I$ into an object representation $f_o$.
%, since it is computationally expensive.
The network continuously observes motion $f_{m,t}$ and hidden state of RNN $h_t$ to parsimoniously trigger the process of computing $f_{o,t+1}$ as follows,
\begin{align}
a_t&=\arg\max_{a}\pi(a\mid(h_t, f_{m,t});W_p)\in\{0,1\}~,&\\
\hat{f}_{o,t+1}&=(1-a_{t})\cdot \hat{f}_{o,t}+a_{t}\cdot f_{o,t+1}(I_{t+1})~,&\label{Eq.sf}\\
g_{t+1}&=\textrm{Emb}(W_{emb},\textrm{con}(f_{m,{t+1}}, \hat{f}_{o,{t+1}}))~,&\label{Eq.ng}
\end{align}
where $a_t$ is the decision of our policy network to trigger ($a_t=1$) or not trigger ($a_t=0$), $W_p$ is the parameters of the policy network, the policy $\pi$ outputs a probability distribution over trigger ($a_t=1$) or non-trigger ($a_t=0$), and $\hat{f}_{o,t+1}$ is the modified object representation. 
As shown in Eq.~\ref{Eq.sf}, when $a_t=1$, the visual observation at frame $t+1$ will be updated ($\hat{f}_{o,t+1}=f_{o,t+1}(I_{t+1})$) with high cost on CNN inference. When $a_t=0$, the previous representation will simply be kept ($\hat{f}_{o,t+1}=\hat{f}_{o,t}$). The modified object representation $\hat{f}_{o,t+1}$ will influence the embedded representation $g_{t+1}$ as shown in Eq.~\ref{Eq.ng}.

\noindent\textbf{Reward.}
We set our reward to encourage less triggered operation ($a=1$) while maintaining correct intention anticipation ($y=y^{\textrm{gt}}$) as shown below.
%It's trade-off between these two reward, frequently trigger  will improve the performance of anticipation, but cost lots of resource to deal with images.
%\noindent\textcolor{red}{reward 1.}\\
%Hence, we design two reward functions to solve this problem.\\
%(1) \textbf{Action Reward.} Each time the action decides to see the image, it will get a negative reward, on the other hand, the reward is positive.\\
%(2) \textbf{Intention Reward.} Since correctly anticipate the intention is our ultimate goal. It'll get a large positive reward when the anticipation is right, otherwise, a large negative reward.
%\noindent\textcolor{red}{reward 2.}\\
%Hence, we design a reward function\\
\begin{eqnarray}
R=\left\{
\begin{array}{l}
p_t(y^{\textrm{gt}})\cdot R^+\cdot (1-\dfrac{n}{T}),~\textrm{ if }~y = y^{\textrm{gt}} \\
p_t(y^{\textrm{gt}})\cdot R^-\cdot \dfrac{n}{T},~~\textrm{ if }~~~y \neq y^{\textrm{gt}} \label{Eq.R}
\end{array}
\right .
\end{eqnarray}
where $y^{\textrm{gt}}$ is the ground truth intention, $y$ is the predicted intention, $n$ is the number of triggered operations in $T$ frames of the video, $p_t$ is the probability of anticipated intention, $R^+$ is a positive reward for correct intention anticipation, and $R^-$ is a negative reward for incorrect intention anticipation. Note that, when the trigger ratio $n/T$ is higher, the positive reward is reduced and the negative reward gets more negative.
%\sunmin{Also, the reward is weighted by the anticipation probability $p_t$.}
%will alert the model when there is a positive motion-trigger decision. When probability is high, that means the motion-trigger decision is confident. It's acceptable if the reward is $R^+$, however, if a confident decision leads to a negative reward, the punishment will also increase.}
\\
% Comment the pre-training part
%\noindent\textbf{Pre-training Procedure.}
%Jointly training the RNN and Policy Network from scratch is challenging since the possible action space (i.e., the unique binary trigger or not pattern) is exponentially large. We follow the intuition of curriculum learning and train RNN with a simple policy as an intermediate step.
%We fix the policy to uniformly skips every $d$ frames as below,
%\begin{eqnarray}
%a_t=\left\{
%\begin{array}{l}
%1,~~t \bmod d=0~,\\
%0,~~otherwise~.
%\end{array}
%\right .
%\end{eqnarray}
%We use the fixed policy to update the embedding function and RNN model. In this way, RNN is guided to handle the modified visual observation in Eq.~\ref{Eq.sf}. After the network converge, we jointly train the whole network (Fig.xx) with an additional policy loss based on the reward.
\noindent\textbf{Policy loss.}
We follow the derivation of policy gradient in~\cite{Williams1992} and define a policy loss function $L^{P}$,
\begin{eqnarray}
L^{P}=-\frac{1}{KT}\sum\limits_{k=1}^K\sum\limits_{t=1}^T \log(\pi(a^k_t\mid(h_t^k,f_{m,t}^k);W_p))\cdot R^k_t~,
\end{eqnarray}
where $\{a_t^k\}_t$ is the $k^{th}$ sequence of trigged patterns sampled from $\pi(\cdot)$, $K$ is the number of sequences, and $T$ is the time when intention reached. $R^k_t$ is the reward of the $k^{th}$ sampled sequence at time $t$ computed from Eq.~\ref{Eq.R}. Please see \highlight{Sec.2 of the} supplementary material for the derivation.

\noindent\textbf{Joint training.}
The whole network (Fig.~\ref{fig.model}) consists of a RNN and a policy network. We randomly initialize the parameters $W_p$ of policy network. The parameters of RNN is initialized by the RNN encoder trained on both representation $f_{o}$ and $f_m$. This initialization enables the training loss to converge faster.
We define the joint loss $L=L^P+\lambda L^A$ for each training example, where $\lambda$ is the weight to balance between two loss. Similar to the standard training procedure in deep learning, we apply stochastic gradient decent using mini-batch to minimize the total joint loss.% with parameters specificed in Sec. xxx.

% \chan{
% There are two objective functions need to be optimized. (1) The cross entropy intention loss with exponential increasing in eq.\ref{eq.exploss} (2) The loss of policy network which defined as
% \begin{eqnarray}
% L^{P}=-\sum\limits_{m=1}^M\sum\limits_{t=1}^T \log(\pi)\cdot r_t
% \end{eqnarray}
% where $\pi$ is the probability of the action, $r_t$ is the reward at frame $t$ which defines in eq.xx.
% } 

\subsection{Learning Representations from Auxiliary Data}\label{sec.CNNs}

Due to the limited daily intention data, we propose to use two auxiliary datasets (object interaction and hand motion) to pre-train two encoders: an object Convolutional Neural Network (CNN) and a hand motion 1D-CNN. In this way, we can learn a suitable representation of object and motion.
%for anticipating intention.

\noindent\textbf{Object CNN.}
%$o=CNN(I)$
It is well-known that ImageNet~\cite{imagenet_cvpr09} pre-trained CNN performs well on classifying a variety of objects. However, Chan et al.~\cite{ChanECCV16} show that images captured by on-wrist camera are significantly different from images in ImageNet. Hence, we propose to collect an auxiliary image dataset with 50 object categories captured by our on-wrist camera, and fine-tuned on Imagenet~\cite{imagenet_cvpr09} pre-trained ResNet-based CNN~\cite{he15deepresidual}.
%We show by experiment that our pre-trained object CNN outperforms the Imagenet pre-trained CNN on intention anticipation.
After the model is pre-trained, we use the model to extract object representation $f_o$ from the last layer before softmax.

\begin{figure}[t!]
\vspace{-5mm}
\begin{center}
\includegraphics[width=0.975\linewidth]{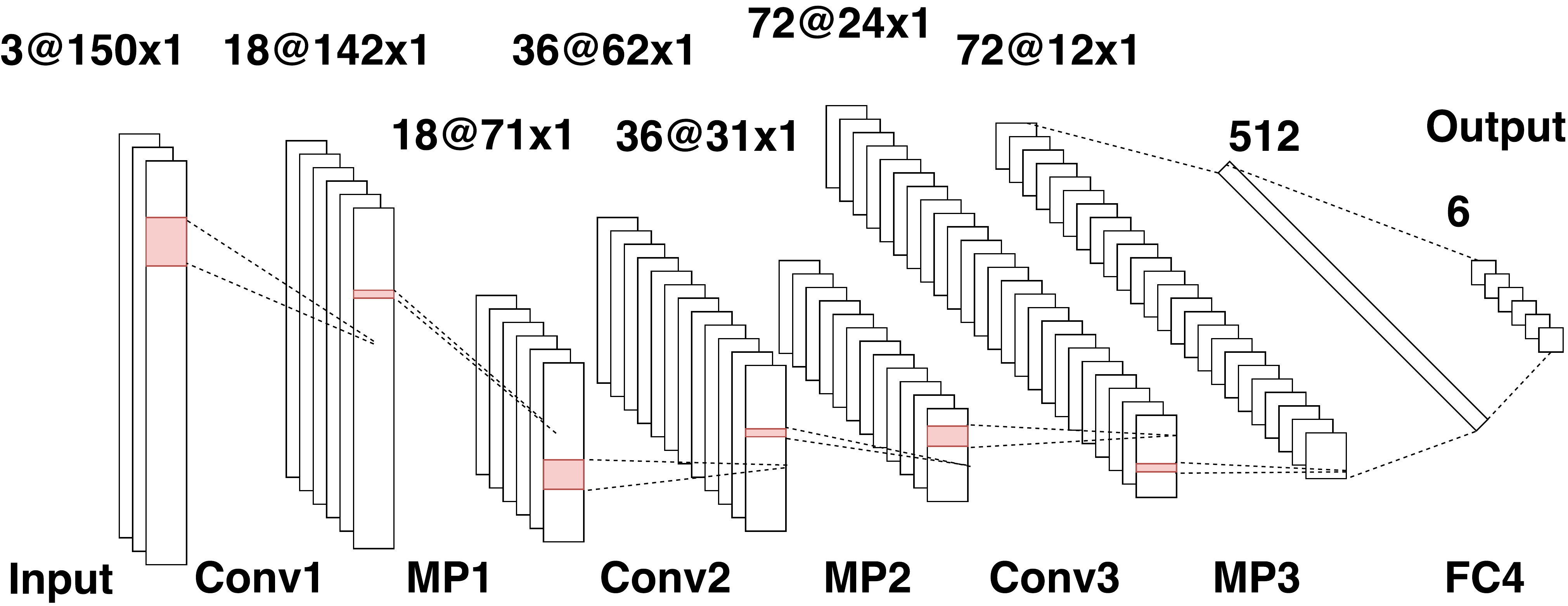}
\end{center}
\caption{\small Illustration of our 1D-CNN pre-trained to classify six motions. Conv, MP, FC stand for Convolution, Max Pooling, and Fully Connected, respectively. $3@150\times 1$ denotes that there are three $150\times 1$ matrices. Since the second dimension is always one, it is a 1D-CNN. Our model has three stacks of Conv+MP layers and a FC layer at the end.}\label{fig.CNN}
\cutfiguredown
\end{figure}

\noindent\textbf{Hand motion 1D-CNN.}
Our accelerometer captures acceleration in three axes ($s\in R^3$) with a sampling rate of 75Hz. We calibrate our sensor so that the acceleration in 3 axes are zero when we placed it on a flat and horizontal surface. We design a 
1D-CNN to classify every 150 samples (2 seconds) into six motions:
lift, pick up, put down, pull, stationary, and walking.
The architecture of our model is shown in Fig.~\ref{fig.CNN}.
Originally, we plan to mimic the model proposed by \cite{Yuqing_2015_IEEESMC}, which is a 3-layer 2D-CNN model with 1 input channel. Considering that there are no stationary properties among three acceleration values for each sample, we adjust the input channel number to 3 and define the 1D-CNN. 
For training the model, we have collected an auxiliary hand motion data with ground truth motions (Sec.~\ref{sec.Dataset}).
%We show in experiment that our 1D-CNN outperforms the 2D-CNN\cite{Yuqing_2015_IEEESMC} on motion classification.
After the model is trained, we use the model to extract motion representation $f_m$ at the FC4 layer (see Fig.~\ref{fig.CNN}).

\subsection{Implementation Details}

\noindent\textbf{Intention anticipation model.}
We design our intention anticipation model to make a prediction in every half second. 
All of our models are trained using a constant learning rate 0.001 and 256 hidden states.

\noindent\textbf{Policy Network.}
Our policy network is a neural network with two hidden layers. For joint training, we set learning rate 0.001, $\lambda$ 0.1 for joint loss. The reward of $R^+$ and $R^-$ are 100 and -100, respectively.

\noindent\textbf{Object CNN.}
Following the setting of~\cite{ChanECCV16}, our object CNN aims at processing 6 fps on NVIDIA TX1. This frame rate is enough for daily actions. Since most of the actions will last a few seconds, it's unnecessary to process at 15 or 30 fps. We take the average over 6 object representations as the input of our model.
Different from~\cite{ChanECCV16}, our on-wrist camera has a fish-eye lens to ensure a wide field-of-view capturing most objects.
For fine-tuning the CNN model on our dataset, we set maximum iterations 20000, step-size 10000, momentum 0.9, every 10000 iteration weight decay 0.1, and learning rate 0.001. We also augment our dataset by horizontal flipping frames.

\noindent\textbf{Hand motion 1D-CNN.}
Motion representation is extracted for a 2-second time segment. Hence, at every second, we process a 2-second time segment overlapped with previous processed time segment for 1 second.
For training from scratch, we set the base learning rate to 0.01 with step-size 4000, momentum 0.9 and weight decay 0.0005. We adjust the base learning rate to 0.001 when fine-tuning.
%our 1D-CNN model.
\cutsectionup
\section{Setting and Datasets}\label{sec.Dataset}
\cutsectiondown
%\highlight{Update the information of Dataset (add User C)!!}

\begin{figure}[t!] 
\vspace{-5mm}
\begin{center}
\includegraphics[width=0.8\linewidth]{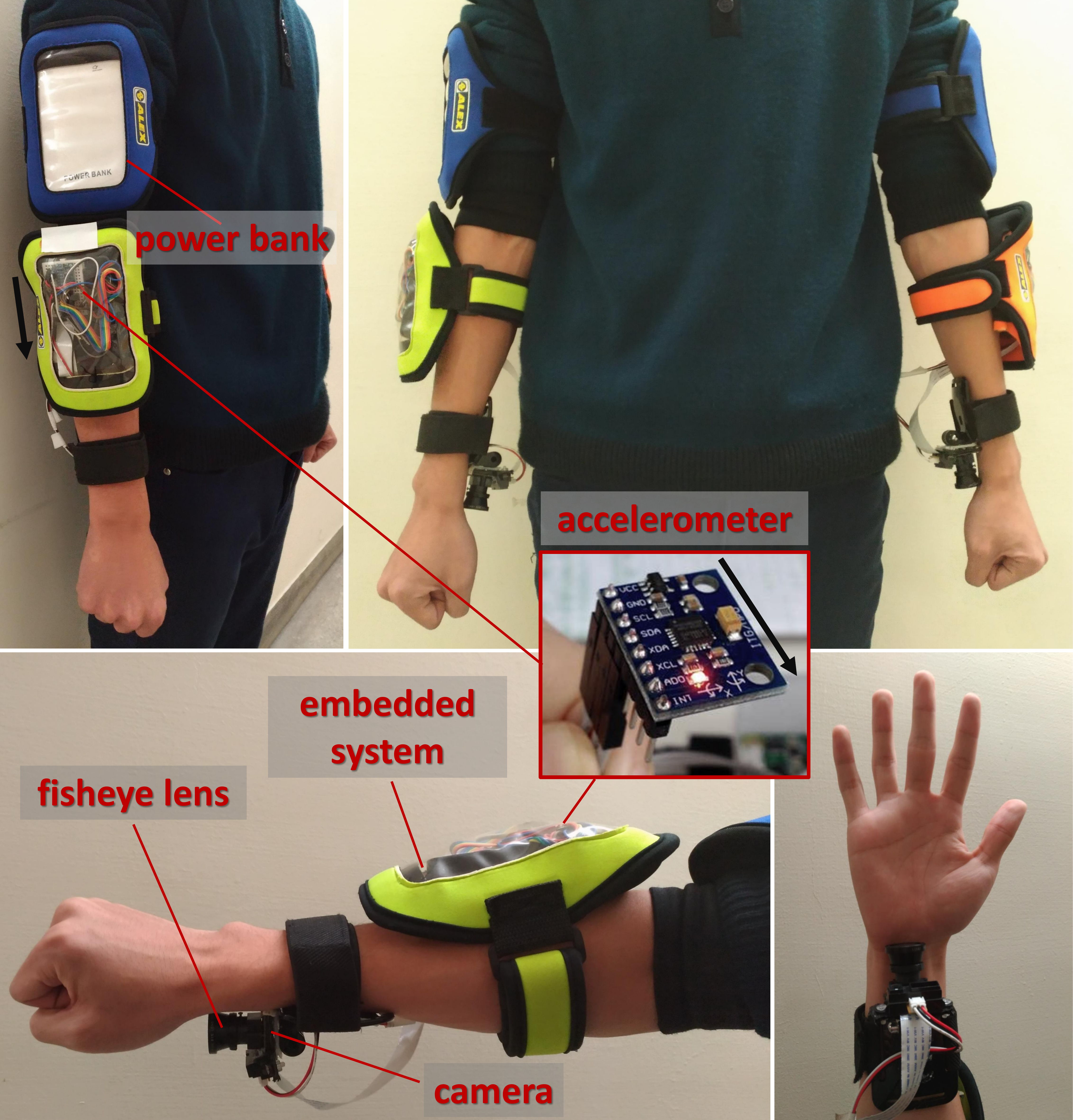}
\end{center}
\vspace{-4mm}
\caption{\small Our on-wrist sensing system. The fisheye camera is below the wrist. The embedded system and motion sensor are on the forearm. Both hands are equipped with the same system.}\label{fig.device}
\vspace{-5pt}
\end{figure}

\begin{figure*}[t!] \vspace{-5mm}
\begin{center}
\includegraphics[width=0.8\linewidth]{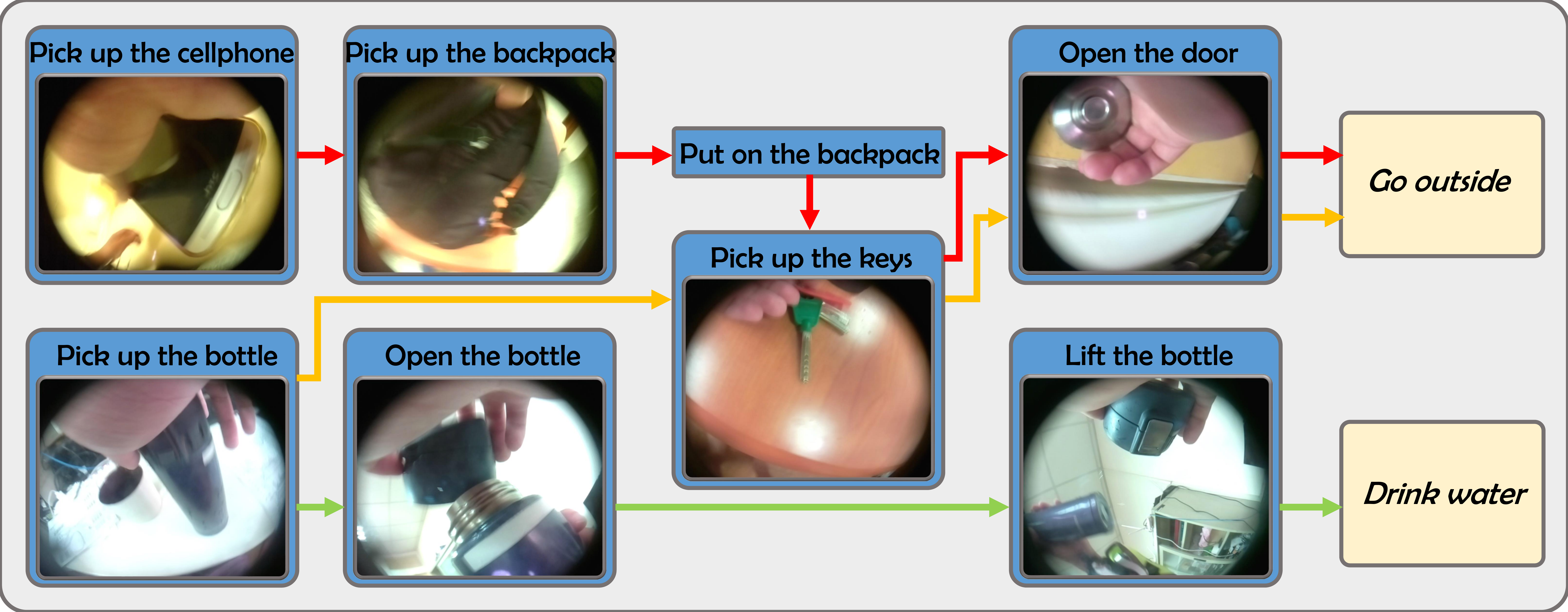}
\end{center}
\vspace{-4mm}
\caption{\small Daily intention dataset. We show examples of two action sequences (red line and yellow line) reaching to the same intention (go outside). In yellow line and green line, we show that the same object (bottle) involves in two intentions (go outside vs. drink water).}\label{fig.ID}
\vspace{-6pt}
\end{figure*}

We introduce our setting of on-wrist sensors and describe details of our datasets.
\cutsubsectionup
\subsection{Setting of On-wrist Sensors}\label{sec.device}
\cutsubsectiondown
Following similar settings in ~\cite{Ohnishi_2016_CVPR,ChanECCV16}, our on-wrist camera\footnote{fisheye lens mounted on noIR camera module with CSI interface.} and accelerometer\footnote{MPU-6050.} are mounted as shown in Fig.~\ref{fig.device}. 
Both camera and accelerometer are secured using velcro. We use the fisheye lens to ensure a wide field-of-view. We list some simple rules to be followed by users. First, the camera is under the arm, toward the palm. Second, the camera should roughly align the center of the wrist. This ensures that camera can easily record the state of the hand. %In this setting, the hand rarely occludes the camera (also observed in~\cite{Ohnishi_2016_CVPR,ChanECCV16}). We will release our system design once our work is accepted.

\subsection{Datasets}
We collect three datasets\footnote{Our dataset and code can be downloaded from \url{http://aliensunmin.github.io/project/intent-anticipate}} for the following purposes.
%\begin{itemize}
(1) Daily Intention Dataset: for training our RNN model to anticipate intention before the intention occurs.
(2) Object Interaction Dataset: for pre-training a better object interaction encoder to recognize common daily object categories.
(3) Hand Motion Dataset:
for pre-training a better motion encoder to recognize common motions.
%\end{itemize}

% . In intention dataset, the user is asked to performed a series of atomic actions in normal daily life, such like pick up the bottle, open the door, and so on. At the end, these actions will refer to a human intention, for example, go outside. Also, we collected a objects dataset which contains some common objects in daily life, and a motion dataset including some general hand motions. We hope these two third-party dataset can be used to make intention anticipation better.

\begin{table}[!h]
\begin{center}
\small
\begin{tabular}[t]{c|ccc}
\hline
User & A & B & C \\
\hline
\# of action sequences & 1540 & 358 & 481\\
avg. per sequence & 9.4 & 2.2 & 2.9\\
\hline
\end{tabular}
\end{center}
\vspace{-3mm}
\caption{\small Statistics of our intention dataset.}\label{Table.Int}
%\cutfiguredown
\normalsize
\vspace{-2mm}
\end{table}

\subsubsection{Daily Intention Dataset}\label{sec.DIdata}
%\cutsubsectiondown
\vspace{-0.05in}
Inspired by Sigurdsson et al.~\cite{Hollywood}, we select $34$ daily intentions such as charge cellphone, go exercise, etc.
Note that each intention is associated with at least one action sequence, and each action consists of a motion and an object (e.g., pick up+wallet). We propose two steps to collect various action sequences fulfilling $34$ daily intentions.

%Next, we collect samples of action sequences and intentions in two stages.

%\chan{\noindent\textbf{Definition.}}\chan{Before we go through the details, we first define something more clearly. This dataset contains 34 daily {\bf intentions}. For each intention it will contains at least one {\bf action sequence}. And each action sequence is consisted of a series of daily {\bf actions} (e.g., pick up the wallet) with its unique order.}

\noindent\textbf{Exploring stage.}
At this stage, we want to observe various ways to fulfill an intention (Fig.~\ref{fig.teaser}).
Hence, we ask a user (referred to as user A) to perform each intention as different as possible. At this step, we observed $164$ unique action sequences.

\noindent\textbf{Generalization stage.}
At this stage, we ask user A and other users (referred to as user B and user C) to follow $164$ action sequences and record multiple samples\footnote{10, 2, 3 times for user A, B, C, respectively} for each action sequence. This setting simulates when an intelligent system needs to serve other users. We show by experiment that our method performs similarly well on three users.

% \noindent\textbf{Scaling up stage.}
% At this stage, we fix the action sequences and focus on scaling up the number of samples of each sequence. Hence, we ask user A to follow $164$ action sequences multiple times. This setting simulates an intelligent system serve specific user for a period of time. At the end, we record $1540$ sample instances with synchronized image and motion sensory data on both hands \sunmin{(Fig.xxx)}. Each action sequence is recorded about $10$ times on average.

In Table~\ref{Table.Int}, We summarize our intention dataset. Note that the number of action sequences recorded by user A is much more than others. Since we will train and validate on user A, selecting the proper hyper-parameters (e.g., design reward function). Next, we'll apply the same setting to the training process of all users, and evaluate the result. This can exam the generalization of our methods.
\highlight{Design of reward function is described in the Sec.3 of the supplementary material.}

\begin{figure} [t!]
\begin{center}
\includegraphics[width=0.975\linewidth]{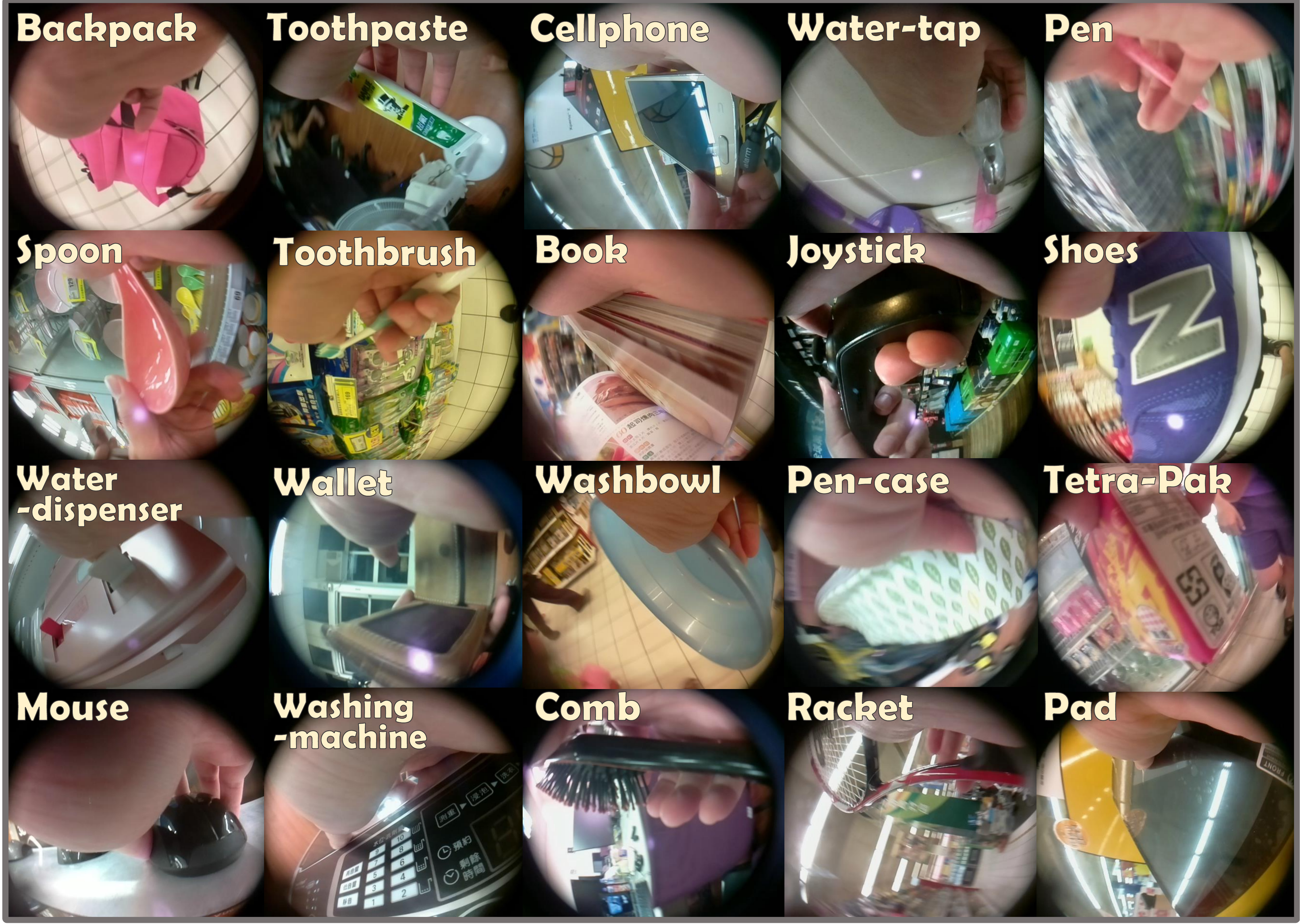}
\end{center}
\vspace{-4mm}
\caption{\small Auxiliary object dataset. Sample images overlaid with their ground truth categories.}\label{fig.auxO}
\vspace{-6mm}
\end{figure} 

\vspace{-2.5mm}
\subsubsection{Object Interaction Dataset.}
\vspace{-2.5mm}
We select $50$\footnote{including a hand-free category.} object categories and collect a set of \gina{$940$} videos corresponding to \gina{$909$} unique object instances\footnote{not counting ``free'' as an instance.}.
Each video records how an object instance is interacted by a user's hand. We sample \gina{$362$} frames on average in each video.
%We ask $xxx$ user in total to collect this dataset.
At the end, we collected an auxiliary dataset consisting of \gina{$340,218$} frames in total to pre-train our object encoder.
Example frames of the dataset are shown in Fig.~\ref{fig.auxO}.
%The background and lighting condition between images in Fig.~\ref{fig.auxO} and Fig.~\ref{fig.ID} are fairly different since we collect the auxiliary dataset mostly in shopping centers rather than home or office space.
%In order to test how this affects the classifier's accuracy, we also collect a smaller dataset in home as testing data.
%50 categories, xxx instances, no intention, xxx users.

\vspace{-3mm}
\subsubsection{Hand Motion Dataset} 
Inspired by \cite{Yuqing_2015_IEEESMC}, we select six motions. We ask eight users to collect $609$ motion sequences from the right hand
and one user to collect $36$ motion sequences from the left hand.
For the right-hand data collected by eight users, we aim at testing cross users generalizability.
For the left-hand data, we aim at testing cross hand generalizability.

% In order to emphasize the discrepancy of motion across users. We ask eight users to collect these data evenly using left and right hand interchangeably.
% \sunmin{[Do we need a figure here?]}

% \begin{figure}
% \begin{center}
% \end{center}
% \caption{Intention Dataset}
% \end{figure}

%\subsection{Object Instance}

%\subsection{Motion Instance}

\vspace{-2mm}
\section{Experiments}\label{sec.Exp}

We first conduct pilot experiments to pre-train object and hand motion encoders. This helps us to select the appropriate encoders.
Next, we conduct experiments for intention anticipation with policy network and evaluate our method in various settings. Finally, we show typical examples to highlight the properties of our method.

\begin{table}[t!]
\begin{center}
\begin{tabular}[t]{c|ccc}
\hline
Model & Training Acc. & Testing Acc.& Speed\\
\hline
VGG-16 & 98.58\% & 77.51\% & 4 fps\\
ResNet-50 & \textbf{99.92}\% & 80.77\% & \textbf{6} fps\\
ResNet-101 & 97.45\% & 82.79\%& 4 fps\\
ResNet-152 & 96.83\% & \textbf{83.09}\% & 3 fps\\
%Model&VGG-16&Res-50&Res-101&Res-152\\
%\hline
%speed&4 fps&\textbf{6} fps& 4 fps&3 fps\\
\hline
\end{tabular}
\end{center}
\vspace{-2mm}
\caption{\small Classification accuracy and processing speed of different models. We highlight the best performance using bold font.}\label{table.oexp}
\end{table}

\subsection{Preliminary Experiments}

%\subsection{Object Pre-training}\label{sec.pipexp}
\noindent\textbf{Object pre-training.}
We evaluate multiple Convolution Neural Network (CNN) architectures on classifying $50$ object categories in our object intention auxiliary dataset. These architectures include VGG-16~\cite{Simonyan14c} and Residual Net (ResNet)~\cite{He2015} with 50, 101, 152-layers. 
We separate the whole dataset into two parts: 80\% of object instances for training and 20\% for testing. The testing accuracy is reported on Table.~\ref{table.oexp}.
Our results show that deeper networks have slightly higher accuracy.
Another critical consideration is the speed on the embedded device. Hence, we report the processed frames per second (fps) on NVIDIA TX1 in the last column of Table.~\ref{table.oexp}. Considering both accuracy and speed, we decide to use ResNet-50 since we designed our system to process at $6$ fps similar to~\cite{ChanECCV16}.

% Convolution neural network (CNN) has achieved great performance on object recogntion nowadays. It has typically had a standard structure, such as stacked convolutional layers (optionally followed by max-pooling) followed by one or more fully-connected layers. Among lots of released CNN models, deep residual network, introduced by Kaiming et al.~\cite{He2015}, is the most powerful model recently. They present a residual learning framework to ease the training of networks, and also won the 1st places on ILSVRC and COCO 2015 competitions. Depends on above reasons, we finetune from 50, 101, 152 layers residual network for our object recognition pretraining, and compare with classic VGG-16 network~\cite{Simonyan14c}.\\
% We use Caffe to train our object recognition model on our own third-party object dataset, which contains 50 categories. The result in Table x shows that the deeper network has higher top-1 accuracy. Our goal is to predict intention via wearable device, so memory size of the model should be chewed over. Due to memory limitation in embedded system nowadays, we finally choose ResNet-50 for our object interaction pretrain model. 

\begin{table}[t!]
\begin{center}
\small
\begin{tabular}[t]{lll}
\hline
Model & Training Acc. & Testing Acc. \\
\hline
1ch-3layer~\cite{Yuqing_2015_IEEESMC} & 100.00\% & 81.41\% \\
3ch-1layer & 100.00\% & 77.21\% \\
3ch-2layer	 & 100.00\% & 78.37\% \\
3ch-3layer	 & 100.00\% & \textbf{83.92}\% \\\hline
left &  & 52.78\% \\
left-\textit{flip} &  & \textbf{83.33}\% \\
\hline
\end{tabular}\normalsize
\end{center}
\vspace{-2mm}
\caption{\small Motion classification accuracy of different models. We highlight best performance using bold font.}\label{table.m}
\cutfiguredown
\end{table}

% Exp Table
\begin{table*}[t!]
\vspace{-5mm}
\centering
\footnotesize
\begin{tabular}{|c|c|c|c|c|c|c|c|c|c|c|c|c|}
\hline
%& \multicolumn{3}{c|}{25\%} & \multicolumn{3}{c|}{50\%} & \multicolumn{3}{c|}{75\%} & \multicolumn{3}{c|}{100\%} \\ \hline
& \multicolumn{4}{c|}{User A} & \multicolumn{4}{c|}{User B} & \multicolumn{4}{c|}{User C} \\ \hline
%\textbf{User}       & A     & B    & C    &  A     & B    & C    & A     & B    & C  & A     & B    & C     \\ \hline
& 25\%     & 50\%    & 75\%    &  100\%    & 25\%     & 50\%    & 75\%    &  100\%    & 25\%     & 50\%    & 75\%    &  100\%    \\ \hline
 
\textbf{Con.}     & 88.41\%     & 90.24\%    & 92.07\%    & 93.29\%     & 90.85\%    & 92.68\%    & 94.51\%     & 95.12\%    &
97.56\%    & 97.56\%     &98.17\%     & 98.17\%    \\ \hline
\textbf{OO}       & 89.63\%     & 92.68\%    & 92.68\%    & 94.51\%     & 91.46\%    & 94.51\%    & 94.51\%     & 95.73\%    & 96.95\%    & 96.95\%     & 98.17\%     & 98.17\%    \\ \hline
\textbf{MO}       & 65.85\%     & 70.73\%    & 75.61\%    & 75.61\%     & 62.20\%    & 66.46\%    & 69.51\%     & 72.56\%    & 71.34\%    & 79.88\%     & 85.37\%     & 87.20\%    \\ \hline\hline
\textbf{Mtr.}     & 86.58\%     & 90.24\%    & 92.07\%    & 92.68\%     & 84.75\%    & 88.41\%    & 88.41\%     & 90.85\%    & 94.51\%    & 96.34\%     & 97.56\%     & 97.56\%    \\ \hline
\textbf{Ratio} & 34.00\%     &32.34\%    & 30.72\%    & 28.42\%     & 31.13\%    & 33.23\%    & 30.88\%     & 29.67\%    & 33.40\%    & 33.88\%     & 30.89\%     & 29.17\%    \\ \hline
\end{tabular}
\vspace{1.5mm}
\caption{\small Intention anticipation comparison. \textbf{OO} stands for object-only observation. \textbf{MO} stands for motion-only observation. \textbf{Con.} stands for concatenating $f_o$ and $f_m$. \textbf{Mtr.} stands for motion-triggered. \textbf{Ratio} stands for triggered ratio.  In the second row, 25\% denotes only the beginning 25\% of the action sequence is observed. All methods are evaluated on A, B, and C users. Note that \textbf{Mtr.} is significantly better than \textbf{MO} and only slightly worse than $\textbf{Con.}$ while processing only about 29\% of the frames.
%We highlight the best performance in all settings with bold fonts.
}
\label{IAexp}
\vspace{1mm}
\end{table*}

\begin{figure*}[t!] 
\vspace{-5pt}
\begin{center}
%\fbox{\rule{0pt}{4in} \rule{0.9\linewidth}{0pt}}
\includegraphics[width=0.95\linewidth]{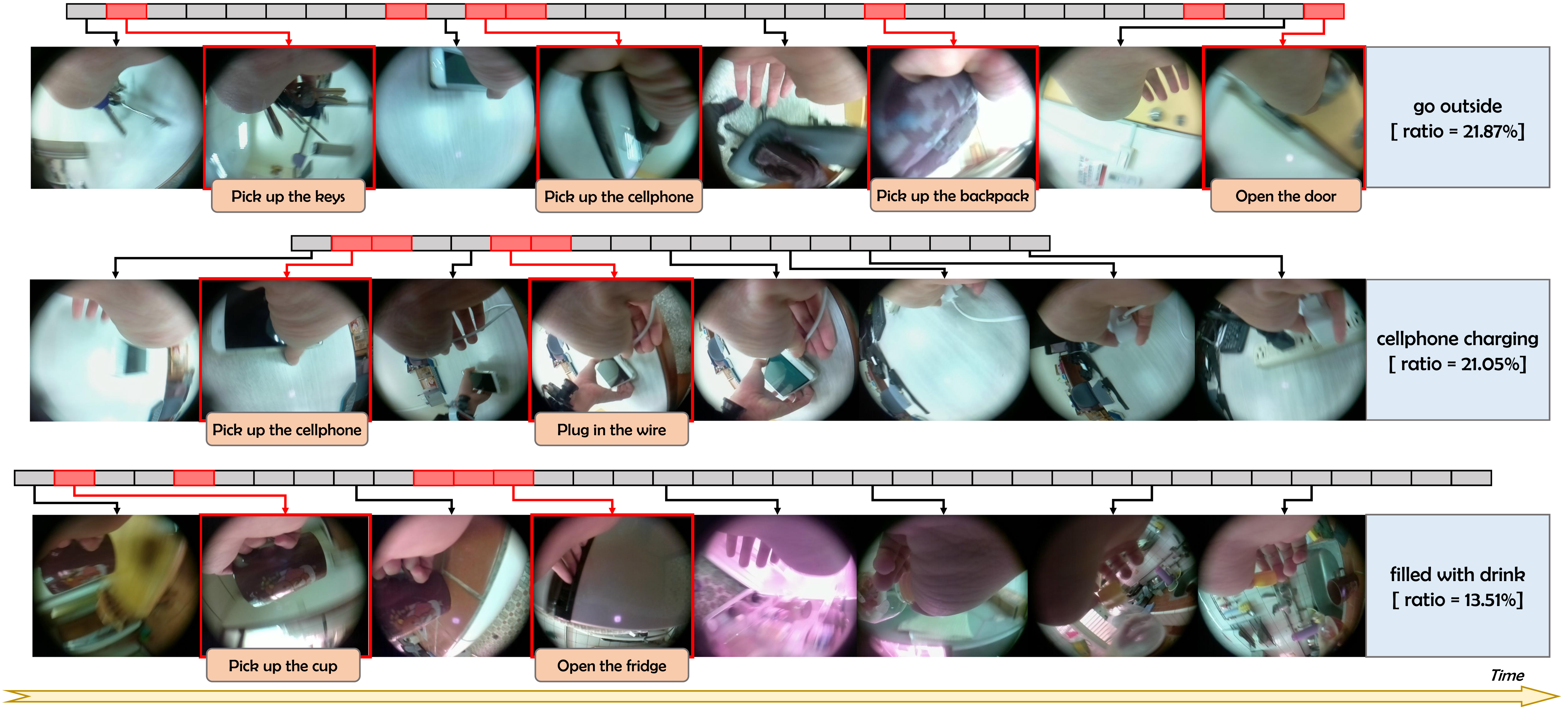}
\end{center}
\vspace{-4mm}
\caption{\small Typical Examples.
In each row, we show an example of our motion-triggered method selecting visual observations. The gray block represents non-triggered frames, and red block represents triggered frames. Each block stands for half second. A few triggered (red boxes) and non-triggered (regular boxes) frames are visualized.
At the end of each example, We show the trigger ratio and the correctly predicted intention. More results are shown in the \highlight{Sec.1 of the} supplementary material.
}\label{fig.te}
%\vspace{-5pt}
\end{figure*}

% adjusting threshold
%\begin{table}
%\begin{center}
%\small
%\begin{tabular}[t]{|c|c|c|c|c|}
%\hline
%\multicolumn{1}{|c|}{\multirow{2}{*}{Threshold}} & \multicolumn{2}{c|}{25\%} & \multicolumn{2}{c|}{100\%} \\
%\cline{2-5}
%& accuracy & Tr.ratio & accuracy & Tr.ratio \\\hline
%$1$ & $66.46\%$ & $00.00\%$ & $63.41\%$ & $00.00\%$ \\
%\hline
%\end{tabular}\normalsize
%\end{center}
%\caption{\small Varying threshold of triggers.}\label{table.th}
%\end{table}

%\subsection{Hand Motion Pre-training}\label{sec.hmexp}

%\noindent\textbf{Hand motion pre-training.}
For hand motion, We describe two experiments to (1) select the best model generalizing across users, and (2) select the pre-processing step generalizing to the left hand.

\noindent\textbf{Generalizing across users.}
Given our dataset collected by eight different users, we conduct a 4-fold cross validation experiment
and report the average accuracy.
We compare a recent deep-learning-based method~\cite{Yuqing_2015_IEEESMC} \chienreplace{(1ch-3layer), our 1-layer (3ch-1layer), 2-layer (3ch-2layer), and 3-layer (3ch-3layer) models }{(1ch\footnote{ch stands for number of input channels}-3layer model) with our 3ch models} trained from scratch in Table.~\ref{table.m}.
The results show that our 3ch-3layer model generalizes the best across different users. %Most importantly, it outperforms \chan{a recent deep learning-based method} proposed by \cite{Yuqing_2015_IEEESMC}.
At the end, we pre-train our 3-layer model on data collected by \cite{WISDM}\footnote{Their data is collected by cellphone's accelerometer while the cellphone is in user's pocket.} to leverage more data. Then, we fine-tune the model on our auxiliary data.% This model (3ch-3layer-finetune) achieves the best accuracy in Table~\ref{table.m}.

\noindent\textbf{Generalizing across hands.}
We propose the following pre-process to generalize our best model (3ch-3layer trained on right hand data) to handle left hand.
We flip the left hand samples by negating all values in one channel (referred to as \textit{flip}). This effectively flips left-hand samples to look similar to right-hand samples.
In the last two rows of Table.~\ref{table.m}, we show the accuracy of left-hand data. Our method with \textit{flip} pre-processing achieves better performance.
In the intention anticipation experiment, we use ``3ch-3layer" and apply \textit{flip} pre-process on left hand.

\subsection{Motion Triggered Intention Anticipation}
For intention anticipation, we evaluate different settings on all three users. In the following, we first introduce our setting variants and the evaluation metric. Then, we compare their performance in different levels of anticipation (e.g., observing only the beginning X percent of the action sequence).
%Finally, we analyze the influence of abandon action sequences.

\noindent\textbf{Setting variants.}\\
%\paragraph{Setting variants.}
(1) Object-only (\textbf{OO}): RNN considering only object representation $f_o$.\\
(2) Motion-only (\textbf{MO}): RNN considering only motion representation $f_m$ .\\
(3) Concatenation (\textbf{Con.}): RNN considering both object $f_o$ and motion $f_m$ representations.\\
(4) Motion-Triggered (\textbf{MTr.}): RNN with policy network, where the input of RNN is determined by the policy network. In this setting, we also report the ratio of triggered moments (referred as to \textbf{Ratio}). The lower the ratio the lower the computation requirement.

\noindent\textbf{Metric.}
%\paragraph{Metric.}
We report the intention prediction accuracy when observing only the beginning $25\%$, $50\%$, $75\%$, or $100\%$ of the action sequence in a video.

%\noindent\textbf{Comparisons} 
Comparisons of different variants on all users  (referred to as user A, B, and C) are shown in Table.~\ref{IAexp}. We summarize our findings below. Object-only (\textbf{OO}) outperforms Motion-only (\textbf{MO}). This proves that object representation is much more influential than motion representation for intention anticipation.
%We also show the performance of concatenating motion and object (\textbf{Con.}).
We also found that concatenating motion and object (\textbf{Con.}) does not consistently outperform Object-only (\textbf{OO}).
Despite the inferior performance of \textbf{MO},
the tendency of \textbf{MO} under different percentage of observation is pretty steady.
This implies that there are still some useful information in the motion representation.
Indeed, \textbf{MTr.} can take advantage of motion observation to reduce the cost of processing visual observation to nearly 29\% while maintaining a high anticipation accuracy ($92.68\%,90.85\%,97.56\%$).
%However, due to the high computationally expensive on processing visual observation, Object-only (\textbf{OO}) is not practical for on-wrist usage. Though the performance of \textbf{MO} is less than \textbf{OO}, the tendency of \textbf{MO} is pretty steady. This shows that there are still some useful information behind the motion data. This implies that our \textbf{MTr.} has the opportunity to selectively take advantage of motion and visual data.}

%\chan{Based on our reward functions, the training procedure is stable. We can see that the action ratio among different users are similar (about xx\% of images will be observed). Comparing to (\textbf{Con.}), despite the model ignores lots of visual information, the accuracy of intention anticipation only decreases xx\%. at most.}

\begin{figure}[h!]
\begin{center}
\includegraphics[width=\linewidth]{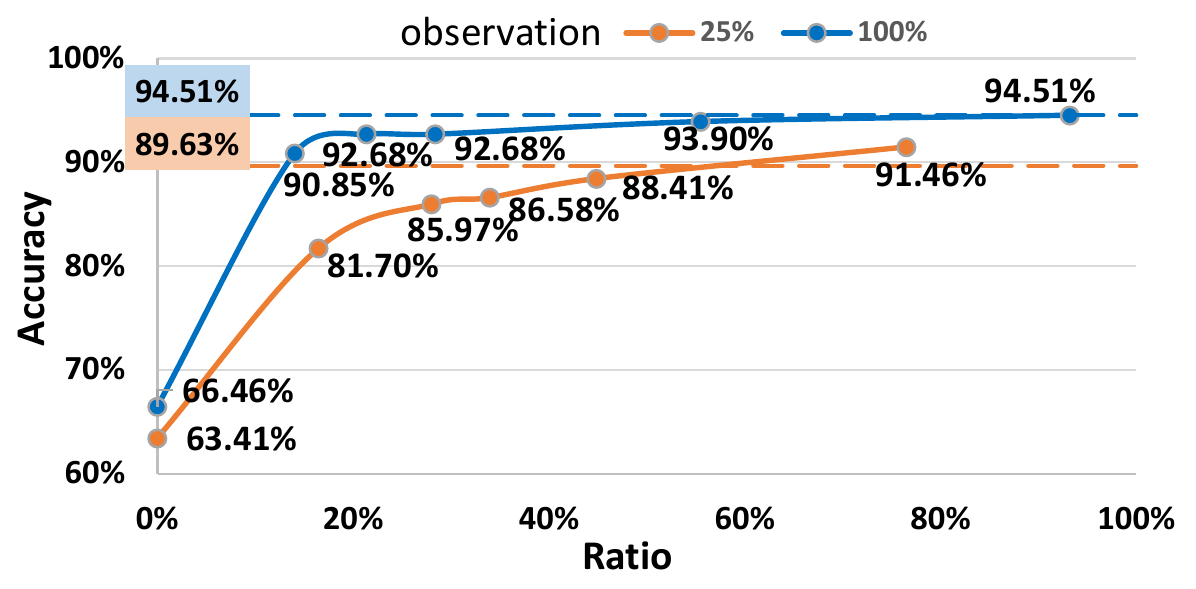}
\end{center}
\vspace{-15pt}
\caption{\small
Anticipation accuracy (vertical axis) of our motion-triggered process on user A for sensing the beginning 25\% (orange solid curves) and 100\% (blue solid curves) of the action sequence. The horizontal axis is the triggered ratio from 0\% (equals to motion-only process) to 100\% (equals to motion-object combined process). We also show the accuracy of object-only process using dash curves.}\label{fig.th}
\vspace{-15pt}
\end{figure}

In Fig.~\ref{fig.th}, we control the ratio of triggered moments and change the anticipation accuracy by adjusting the threshold of motion triggers. 
The results show that increasing the ratio of triggered moments leads to higher accuracy on intention anticipation.
Most interesting, the accuracy only decrease slightly when the ratio is larger than 20\%.
Note that the default threshold is 0.5, which means the policy will decide to trigger when the probability of trigger is larger than non-trigger.
\highlight{Some quantitative results are described in Sec.4 of the supplementary material.}
\vspace{-2pt}
\subsection{Typical Examples}
\vspace{-5pt}
We show typical examples in Fig.~\ref{fig.te}. In the first example, our Policy Network (PN) efficiently peeks at various objects (e.g., keys, cellphone, backpack, etc.). In other examples, PN no longer triggers after some early peeks. Specifically, in the second example, once the cellphone is observed and the wire is plugged in, PN is confident enough to anticipate cellphone charging without any further triggered operation.

%The third example shows a similar situation, after picking up the cup, and opening the fridge. The clues for intention anticipation are enough, and the policy no more trigger.}
\vspace{-12pt}
%We show typical examples in Fig.~\ref{sec.te}. \chienreplace{The first two examples show our method anticipate intention correctly even the action sequences are different. The other two examples show that our method can anticipate correct intention even with only $25\%$ or $50\%$ of the video \chienreplace{sequences}{frames}.}
%{The first example (open air conditioner) shows that at the early stage of the video, the model may be confused by other similar intentions since both ``go into the room" and ``make drink"\footnote{a user open the door and go to press the water-dispenser's button.} share some actions (e.g. open the door). 
%However, the confidence in these predictions are not significantly high. As the model observes more actions, it anticipates the correct intention with more confidence. The second example also shows that the anticipation become more accurate as more video frames are observed.The last example shows an instance that Ours-\textbf{Con.} outperforms Ours-\textbf{Con.}-\textbf{w/o Aux.}, where \textbf{w/o Aux.} makes subtle mistakes due to inferior representation.}

%\noindent\

\section{Conclusion}\label{sec.Con}
\vspace{-5pt}
We propose an on-wrist motion triggered sensing system for anticipating daily intentions.
The core of the system is a novel RNN and policy networks jointly trained using policy gradient and cross-entropy loss to anticipate intention as early as possible. On our newly collected daily intention dataset with three users, our method achieves impressive anticipation accuracy while processing only 29\% of the visual observation. In the future, we would like to develop an on-line learning based method for intention anticipation in the wild.

\section{Acknowledgement}
\cutsectiondown
We thank MOST 104-2221-E-007-089-MY2 and MediaTek for their support.

{\small
\bibliographystyle{ieee}
\bibliography{egbib}
}

\end{document}